\documentclass{article}
\usepackage{arxiv}

\usepackage[utf8]{inputenc} 
\usepackage[T1]{fontenc}    
\usepackage[bookmarks=false]{hyperref}       
\usepackage{url}            
\usepackage{booktabs}       
\usepackage{amsfonts}       
\usepackage{nicefrac}       
\usepackage{microtype}      
\usepackage{lipsum}		
\usepackage{graphicx}
\usepackage{natbib}
\usepackage{doi}
\usepackage{xcolor}
\usepackage{graphicx}
\usepackage{booktabs}
\usepackage{multirow}
\usepackage{amsmath} 
\usepackage{cleveref}
\usepackage{xcolor}
\usepackage{caption}
\usepackage{float}
\usepackage{subcaption}

\title{Improving trajectory continuity in drone-based crowd monitoring using a set of minimal-cost techniques and deep discriminative correlation filters}

\date{} 					

\author{ \href{https://orcid.org/0000-0003-1601-6560}{\includegraphics[scale=0.06]{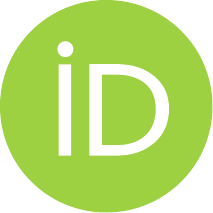}\hspace{1mm}Bartosz Ptak} \\
	Institute of Robotics and Machine Intelligence\\
        Poznań University of Technology\\
	Piotrowo 3A, Poznań, 60-965, Poland \\
	\texttt{bartosz.ptak@doctorate.put.poznan.pl} \\
	\And
	\href{https://orcid.org/0000-0001-6483-2357}{\includegraphics[scale=0.06]{orcid.pdf}\hspace{1mm}Marek Kraft}\thanks{Corresponding author} \\
	Institute of Robotics and Machine Intelligence\\
        Poznań University of Technology\\
	Piotrowo 3A, Poznań, 60-965, Poland \\
	\texttt{marek.kraft@put.poznan.pl}  \\
}

\renewcommand{\shorttitle}{Preprint submitted to Expert Systems with Applications}

\hypersetup{
pdftitle={A template for the arxiv style},
pdfsubject={q-bio.NC, q-bio.QM},
pdfauthor={David S.~Hippocampus, Elias D.~Striatum},
pdfkeywords={First keyword, Second keyword, More},
}

\begin{document}
\maketitle

\begin{abstract}
Drone-based crowd monitoring is the key technology for applications in surveillance, public safety, and event management. However, maintaining tracking continuity and consistency remains a significant challenge. Traditional detection-assignment tracking methods struggle with false positives, false negatives, and frequent identity switches, leading to degraded counting accuracy and making in-depth analysis impossible. This paper introduces a point-oriented online tracking algorithm that improves trajectory continuity and counting reliability in drone-based crowd monitoring. Our method builds on the Simple Online and Real-time Tracking (SORT) framework, replacing the original bounding-box assignment with a point-distance metric. The algorithm is enhanced with three cost-effective techniques: camera motion compensation, altitude-aware assignment, and classification-based trajectory validation. Further, Deep Discriminative Correlation Filters (DDCF) that re-use spatial feature maps from localisation algorithms for increased computational efficiency through neural network resource sharing are integrated to refine object tracking by reducing noise and handling missed detections. The proposed method is evaluated on the DroneCrowd and newly shared UP-COUNT-TRACK datasets, demonstrating substantial improvements in tracking metrics, reducing counting errors to 23\% and 15\%, respectively. The results also indicate a significant reduction of identity switches while maintaining high tracking accuracy, outperforming baseline online trackers and even an offline greedy optimisation method.
\end{abstract}


\section{Introduction}

Drone-based crowd monitoring has emerged as a critical tool in various applications, including surveillance, public safety, and event management~(\cite{emimi2023current}). Drones' ability to provide a bird's-eye view, combined with their mobility and flexibility, offers significant advantages over traditional monitoring systems. Unlike fixed surveillance cameras, drones can cover large, dynamic areas and adapt their position in real time, making them ideal for monitoring large crowds at events, protests, or public gatherings.

\begin{figure}[ht!]
\centering
\includegraphics[width=\textwidth]{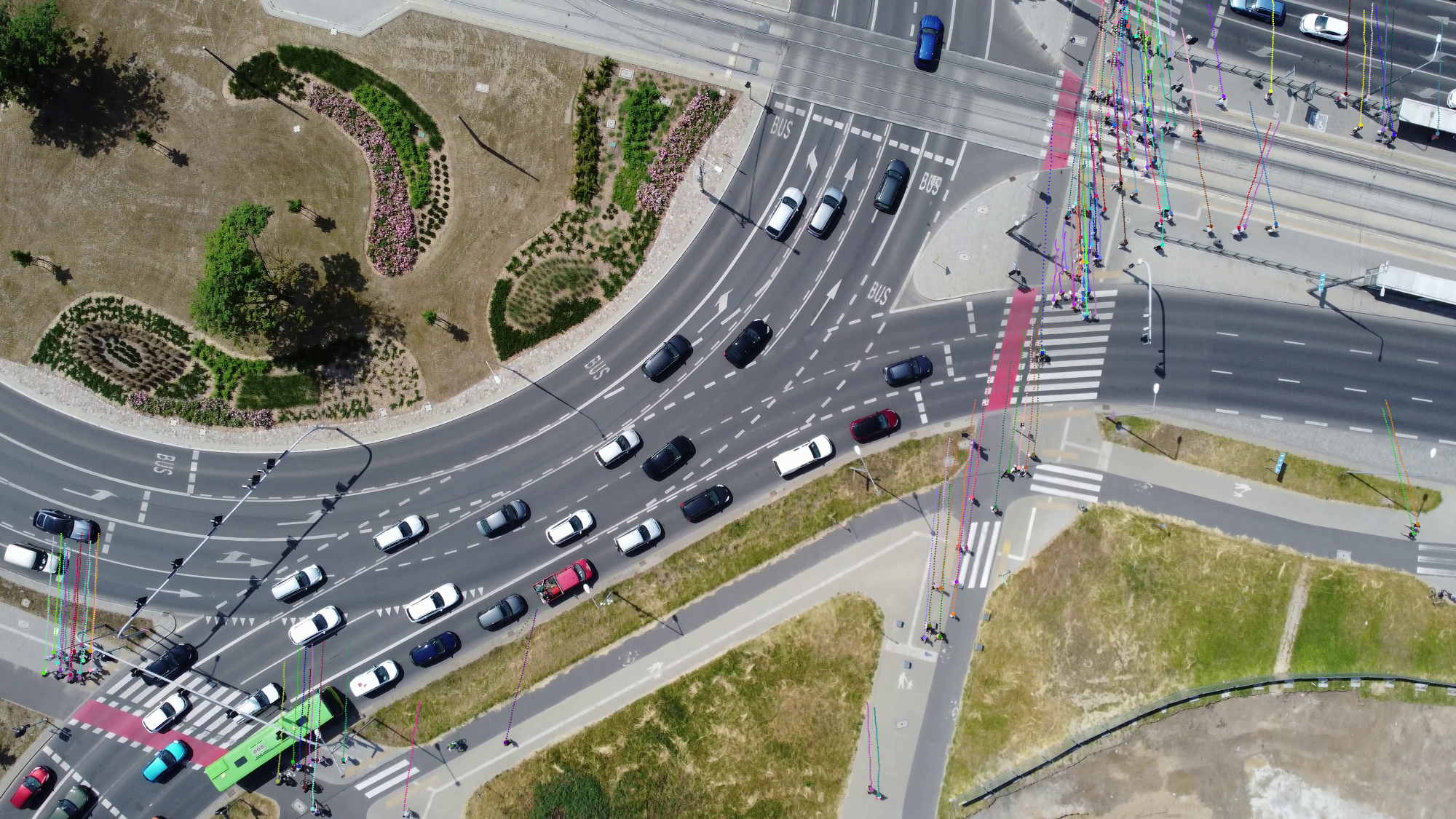}
\caption{An example frame demonstrates the task of localising and tracking tiny objects. Marked trajectories represent the objects' current positions and their movement history over the past two seconds.}
\label{fig:examples:tiny_size}
\end{figure}

Despite its advantages, drone-based crowd monitoring faces several challenges, including dynamic environmental factors, significant crowd occlusions, and the difficulty of avoiding individual re-counting. Although object detection and localisation algorithms have been intensively developed, their performance remains insufficient for maintaining accurate tracking using default detection-assignment-based methods. Ensuring trajectory continuity, which is crucial for effective monitoring and counting, often requires more advanced approaches. In particular, because drone-based crowd object localisation is typically characterised by both false negative and false positive detections. The small size and shape of individuals usually make detection challenging, leading to missing positions in sequential frames, interrupting trajectories, and increasing counting errors. Conversely, environmental objects are frequently misclassified as crowd members, resulting in redundant trajectories and degrading performance metrics, mainly when trajectory continuity improvements are applied. Addressing these challenges necessitates the development of dedicated solutions tailored to the specific requirements of this task, especially for extremely tiny object sizes (less than a few pixels). Fig.~\ref{fig:examples:tiny_size} presents an example frame demonstrating the visual challenges of locating and tracking tiny objects from a drone perspective.

In this article, we propose a point-oriented tracking algorithm designed to enhance the drone-based crowd-counting task that significantly improves metrics and reduces label switches of individual objects by enhancing trajectory continuity and reducing false positive trajectories. Existing people's localisation methods often produce predictions polluted by false positives and false negatives, which becomes particularly problematic during object tracking across video frames. Building on the Simple Online and Real-Time Tracking (SORT) algorithm~(\cite{bewley2016simple}), which fails when dealing with very small and crowded objects, our approach adapts it by replacing the intersection-over-union (IoU) assignment metric with a point-distance metric to measure distances between object coordinates. The method was further improved with three cost-effective enhancements: camera motion compensation to mitigate movement noise, merging of drone altitudes for dynamic assignment, and an additional classification step to confirm trajectory validity. 

Additionally, Deep Discriminative Correlation Filters (DDCFs) are integrated into the tracking pipeline, improving trajectory continuity and consistency by effectively handling incorrect and missed detections. In comparison to existing algorithms, which employ dedicated neural networks as feature extractors, the proposed filters incorporate deep learning features generated by the localisation method that provides the objects' positions in the frame. It ensures robust object tracking while maintaining computational efficiency. The proposed method operates as an online tracker, processing frames sequentially, making it suitable for real-world applications and practical scenarios. The algorithm's accuracy is evaluated using the DroneCrowd~(\cite{wen2021detection}) and the newly introduced UP-COUNT-TRACK datasets.

The principal contributions of this work are as follows\footnote{The code repository and dataset are available: \url{https://up-count.github.io/tracking}.}:
\begin{itemize}
    \item Introduction of a novel point-oriented tracking approach built on the SORT algorithm, replacing the bounding-box assignment metric with a point-distance metric for measuring object coordinate distances, addressing object localisation results. 

    \item Integration of three low-cost improvements to enhance tracking robustness: camera motion compensation to reduce movement noise, dynamic assignment based on drone altitude to align thresholds dynamically, and a classification step to verify trajectory validity before its confirmation.

    \item Proposing a trajectory continuity enhancement method using Deep Discriminative Correlation Filters, which leverage deep learning feature maps from the localisation method to improve tracking consistency while maintaining computational efficiency.

    \item Extending the previously published UP-COUNT (\cite{ptak-upcount}) dataset by introducing the UP-COUNT-TRACK dataset, which comprises a total of 33,751 frames, 3,807 trajectories, and 1,360,547 labelled instances, enabling the evaluation of real-world tracking scenarios.

    \item Evaluate the proposed methods on the UP-COUNT-TRACK and DroneCrowd datasets, comparing performance against the offline global-optimal greedy method to demonstrate tracking accuracy and stability improvements.
\end{itemize}

\section{Related work}

\subsection{Drone-based people localisation methods}

STNNet~(\cite{wen2021detection}) is designed for drone-based tiny object localisation in crowded scenes, featuring a localisation subnetwork that generates object proposals across pixels, including a classification branch that evaluates the likelihood of each proposal and a regression branch that refines their positions. MFA~(\cite{asanomi2023multi}) employs two techniques for feature map generation: heatmap estimation, using an encoder-decoder network to predict object positions based on heatmap peaks, and the Motion and Position Map (MPM), which captures both positions and movement directions by analysing sequential frames and behavioural patterns. The Dot Localisation method~(\cite{ptak-upcount}) addresses the challenge of detecting tiny objects in high-resolution images. It leverages Pixel Distill, a technique that enhances the processing of high-definition images by extracting spatial information from individual pixels, enabling state-of-the-art performance. These methods provide only pure detections for each video frame and require additional assignment steps to perform tracking.

\subsection{Drone-based crowd object tracking}

Most drone-based crowd-tracking methods prioritise the localisation stage, often overlooking the tracking stage, which is equally important for ensuring trajectory consistency and minimising counting errors. Recent state-of-the-art approaches, such as STNNet~(\cite{wen2021detection}) and MFA~(\cite{asanomi2023multi}), employ the Globally-Optimal Greedy (GOG) algorithm~(\cite{pirsiavash2011globally}). GOG is a multi-object tracking method based on a minimal-cost flow framework, enabling it to handle large input sequences and manage long-term occlusions. These features make it particularly useful for dense-object tracking and long sequences. However, as a globally optimised, offline method, it requires access to the entire sequence data, limiting its applicability in real-world tracking scenarios.

\subsection{Visual correlation filters}

The visual tracking task involves estimating an object's trajectory in a video. One of the most applied methods for this task is the Correlation Filter, which has shown outstanding results on benchmarks~(\cite{wu2013online,kristan2015visual}). While the general mechanism of operation remains largely consistent, numerous feature extraction approaches have been proposed. These include multi-channel feature processing using classical features~(\cite{galoogahi2013multi}), non-linear circulant data regression~(\cite{henriques2014high}), and the use of multi-scale colour attributes~(\cite{danelljan2014adaptive}). Subsequent research has shifted focus not only toward developing improved features but also their fusion. For instance, the integration of Histogram of Gradient~(\cite{felzenszwalb2009object}) and colour-naming features~(\cite{van2009learning}) has been combined through scale adaptation techniques~(\cite{li2015scale}). Building upon these same features, a more advanced approach based on channel and spatial reliability has been proposed~(\cite{lukezic2017discriminative}). This method estimates spatial reliability maps by solving graph labelling problems and adapts the filter support to the most reliable parts of the object for tracking.

Although numerous hand-crafted feature-based methods achieved significant results in visual tracking, they were eventually outperformed by convolution-based approaches. One of the earliest methods employed a denoising autoencoder network to generate convolutional filters~(\cite{wang2013learning}), enabling the integration of trainable features into Correlation Filter tracking. With the advent of deep learning, more advanced solutions were developed. For example, a method using stacked hierarchical features embedded by a two-layer convolutional neural network was introduced~(\cite{wang2015video}). Another approach leveraged deep convolutional neural networks to extract feature maps that capture both spatial detail and semantic information~(\cite{ma2015hierarchical}). Similarly, deep features combined with activation layers were utilised as feature maps for tracking~(\cite{danelljan2015convolutional}). 

A breakthrough in visual object tracking with Correlation Filters occurred with the introduction of continuous operators for confidence map generation~(\cite{danelljan2016beyond}). This method generates multi-resolution deep feature maps in a continuous domain, enhancing classical Discriminative Correlation Filters and enabling precise sub-pixel localisation. Building on this advancement, the Efficient Convolution Operators (ECO) algorithm was proposed~(\cite{danelljan2017eco}). The authors introduced a factorised convolution operator that significantly reduces the number of model parameters and developed an efficient model update strategy. This strategy employs a sparse updating scheme that triggers optimisation only when sufficient change in the objective function is detected, thereby improving both tracking speed and robustness.

\section{UP-COUNT-TRACK dataset}\label{sec:dataset}

\begin{figure}[h!]
\centering
\includegraphics[width=\textwidth]{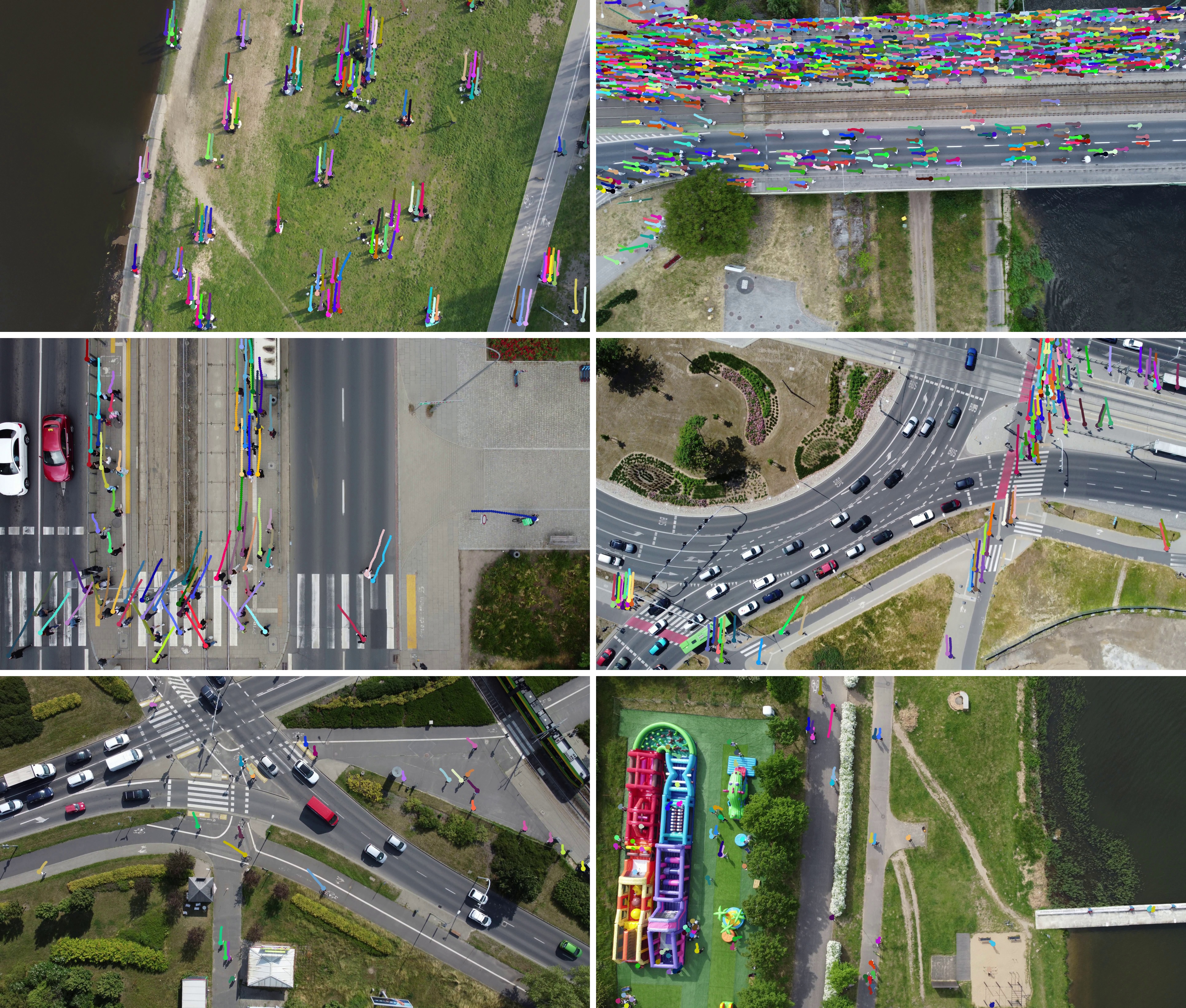}
\caption{Example frames with annotated trajectories illustrate the dataset's diverse recording environments, varying flight altitudes, and dynamic drone movements.}
\label{fig:dataset:examples}
\end{figure} 

In this study, we extend the previously introduced UP-COUNT dataset (\cite{ptak-upcount}) by providing tracking annotations to the test subset. Overall, UP-COUNT comprises 202 unique sequences of sparse drone footage, with frames sampled at one-second intervals from original videos, focusing on point-based people localisation. The dataset is divided into training, validation, and test subsets, containing 141, 30, and 31 sequences, respectively. The new UP-COUNT-TRACK dataset includes tracking labels for all 31 sequences in the test subset, considering each frame of full-time video recordings. These sequences vary in length, ranging from 361 to 2,541 frames, with an average of 1,088.7 frames and a total of 33,751 frames. To facilitate tracking evaluation, each person’s position was manually annotated across all frames, maintaining consistent identification labels throughout each scene. The number of trajectories per sequence ranges from 13 to 1,182, with an average of 122.8 trajectories. Overall, the dataset contains 3,807 trajectories and 1,360,547 labelled instances, highlighting its extensive scale. The continuity and correctness of the labels were reviewed and validated by independent evaluators, separate from the original annotators. Additionally, the dataset’s diverse recording environments, varying flight altitudes, and dynamic drone movements create a challenging benchmark that reflects real-world scenarios and potential urban applications. Moreover, supplemental metadata was stored, providing GPS coordinates and flight altitude, enabling the development of drone-based prototypes of applications. Example frames with annotated trajectories are shown in Fig.~\ref{fig:dataset:examples}.

\section{Method}

This section provides a detailed description of the proposed method, including the people localisation approach used in this research (Section~\ref{sec:method:loc}), the baseline tracking method and its configuration (Section~\ref{sec:method:track}), and the techniques applied to enhance crowd tracking (Section~\ref{sec:method:techniques}). Section~\ref{sec:method:dcf} elaborates on the use of Deep Discriminative Correlation Filters for improving trajectory consistency.

\subsection{People localisation method}\label{sec:method:loc}

The Dot Localisation method~(\cite{ptak-upcount}) is employed to generate person detections for each frame. This state-of-the-art approach for crowd localisation in drone-based images provides a robust foundation for our tracking method. It achieves an L-AP@10 of 57.06 on the DroneCrowd dataset and 75.46 on the UP-COUNT dataset, highlighting opportunities for improving tracking by addressing missing individuals. Additionally, this U-type method outputs spatial features at the decoder stage, which are utilised in our deep discriminative correlation filters (see Section~\ref{sec:method:dcf}), enabling efficient feature tracking.

\subsection{Baseline tracking method}\label{sec:method:track}
The proposed point-oriented tracking method is based on the Simple Online and Realtime Tracking (SORT) algorithm~(\cite{bewley2016simple}), which has been widely enhanced and applied for object tracking applications~(\cite{miranda2023fruit, mirzaei2023small}). SORT employs a Kalman filter for modelling object motion and the Hungarian algorithm for object assignment using a defined comparison metric. Despite its limitations, SORT serves as a reliable and efficient foundation, providing the core tracking logic for our method.

A distance-based assignment method was used to associate objects with known trajectories. This approach computes the Euclidean distance between the coordinates of each detected point marking the object's location and the predicted coordinates of the corresponding trajectory object. The distances are measured in pixels. Matching correctness is determined within a circular region centred on the predicted coordinate. A detected point is considered a correct match if it lies within this circle and the distance to the predicted coordinate is less than the circle's radius. By default, the radius is set to ten pixels, which is half of the twenty-pixel object size used in DroneCrowd.

Adequately adjusted operating parameters are essential for successful tracking algorithms. One key parameter is the minimum number of hits required for a trajectory to be marked as confirmed, representing the minimum trajectory length. Another one is the maximum trajectory age, which specifies how long a missed trajectory is kept in memory before being removed. In our setup, these values are set to 30 and 60 frames, corresponding to approximately 1 and 2 seconds, respectively. These settings make the algorithm robust enough to reduce both false positives and false negative trajectories for long and dynamic sequences.


\subsection{Minimal-cost enhancing techniques}\label{sec:method:techniques}

This section discusses methods that enhance crowd object tracking while ensuring minimal impact on overall performance.

\subsubsection{Camera motion compensation}

When a drone moves during video recording, algorithms must address noise caused by camera motion. The simultaneous movement of objects and the camera complicates accurate trajectory estimation. To mitigate this, we adopted the Camera Motion Compensation module from BoT-SORT~(\cite{aharon2206bot}). This module calculates sparse optical flow to estimate the average translation and rotation between consecutive frames. The process involves identifying sparse corners in both frames and matching them using the Lucas-Kanade method~(\cite{bouguet2001pyramidal}). The matched keypoints are then used to compute an affine transformation matrix between frames. These values are applied to the Kalman filter's state vector and noise matrix, enabling object localisation to be adjusted with the transformation matrix. When acquired images have enough visible features to extract the difference between sequential frames, this approach effectively removes the influence of camera movement from the modelled object's motion, improving the accuracy of object trajectory tracking.

\subsubsection{Drone's sensor altitude information}

Modern aerial platforms are equipped with various sensors, including barometers, accelerometers, gyroscopes, and satellite navigation systems. Despite their availability, data from these sensors are rarely utilised in computer vision-based remote sensing algorithms. Flight altitude above ground level, a valuable metadata source, is often overlooked despite its potential for enhancing algorithm performance (\cite{wilinskielevating}), especially when the drone's camera is precisely calibrated.

The UP-COUNT-TRACK dataset provides altitude information for each frame, with values ranging from 28.9 to 100.7 meters, enabling precise utilisation of this metadata. In contrast, the DroneCrowd dataset lacks direct altitude information. Instead, sequences are divided by object size and categorised into small (sequences: 11, 15, 16, 22, 23, 24, 34, 35, 42, 43, 44, 61, 62, 63, 65, 69, 70, 74) and large (sequences: 17, 18, 75, 82, 88, 95, 101, 103, 105, 108, 111, 112) subsets. Based on this categorisation, we assumed that small objects were recorded at higher altitudes (adopting 100 meters) and large objects at lower altitudes (adopting 50 meters). These assumptions allow us to emulate flight height, even though this dataset does not contain this metadata.

Considering flight altitude, we propose utilising this value to calculate a dynamic threshold for assignment correctness, defined as:
\begin{equation}
    \text{Tr}=max(10, \frac{100}{\text{altitude}}\cdot 10)
\end{equation}
where 10 pixels is half of the default evaluation object size formed in~(\cite{wen2021detection}).
This dynamic threshold allows the tracking algorithm to adapt to variations in object size and inter-object distances. At lower altitudes, where objects occupy more pixels, a larger separation is required for assignment. Conversely, at higher altitudes, where objects are represented by fewer pixels, a smaller range is used to minimise trajectory label switches.

\subsubsection{Additional classification steps to reduce false positives}

In most tracking algorithms, before a trajectory is confirmed, it must persist for a minimum duration specified in the algorithm parameters. This requirement reduces the occurrence of very short trajectories and minimises false positives. In this research, we extend the confirmation process by introducing an additional classification step. If the counter of trajectory presence ($N_{active}$) exceeds a threshold based on the minimal trajectory duration ($N_{thresh}$), defined as:
\begin{equation}
    N_{active} >= (N_{thresh} - 3)
\end{equation}
the object's surroundings are classified until the trajectory is either terminated or confirmed as valid. The trajectory is marked as confirmed if it reaches minimal trajectory duration ($N_{thresh}$) and the average probability of classified localisation surroundings containing objects is higher than 80\%. The value was determined experimentally.

\begin{figure}[h!]
\centering
\includegraphics[width=\textwidth]{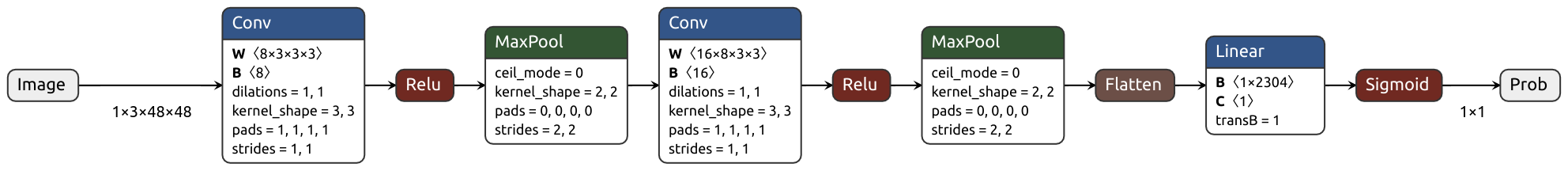}
\caption{Architecture of a simple convolutional network designed to classify if a region of interest contains a person.}
\label{fig:simple_network}
\end{figure} 

A minimal neural network is employed as the classification model, with its architecture shown in Fig.~\ref{fig:simple_network}. The network operates on regions of interest extracted from the colour image based on the centre of the object and dynamic assignment range threshold. These fragments are resized to $48\times48$ pixels before being processed. The network features two blocks, each consisting of a convolutional operation, ReLU activation, and max pooling, to extract image features. The extracted features are then processed through a linear transformation and normalised to the zero-one range using a Sigmoid activation function.

The training dataset is constructed using known object positions in images, along with randomly selected negative examples. Negative samples, predominantly containing background, are selected to balance the positive samples. The selection algorithm maximises sample diversity by ensuring sparse positions within the image. The training process employs binary cross-entropy loss and the Adam optimiser~(\cite{kingma2014adam}) with a learning rate of $5 \cdot 10^{-4}$. Model performance metrics and results are detailed in Section~\ref{sec:cls:results}.

\subsection{Deep Discriminative Correlation Filters}\label{sec:method:dcf}

\begin{figure}[h!]
\centering
\includegraphics[width=\textwidth]{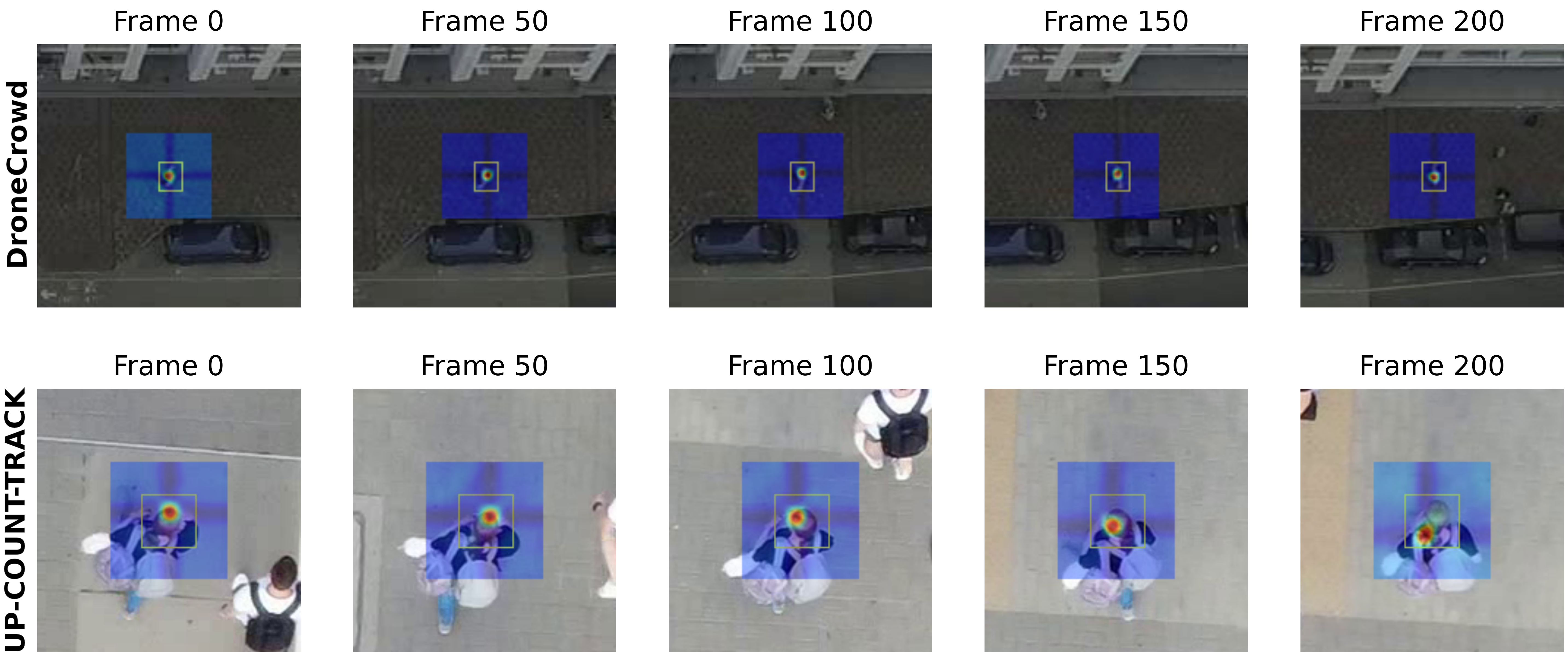}
\caption{Usage of Deep Discriminative Correlation Filters in drone-based people tracking for sample images. The heatmap indicates the locations with the highest correlation within the analysed image area.}
\label{fig:deep:dcf}
\end{figure}

In the proposed tracking algorithm, we adapt the Deep Discriminative Correlation Filters (DDCF) to follow the trajectories of known objects despite missing detections. Once a known and confirmed trajectory is not matched with a new detection in a current frame, the filter is initialised for the object based on the last-measured localisation and visual features. Specifically, our method is built upon the Efficient Convolution Operators (ECO) algorithm~(\cite{danelljan2017eco}) that provides efficient object re-localisation by the fast computation of correlations in the Fourier domain. By combining deep learning-based visual features (Section~\ref{sec:deep_visual_features}) extracted from the localisation model with the adapted DDCF (see Section~\ref{sec:alltogether} for details how the algorithm is constructed), the method effectively enhances trajectory consistency. A visual interpretation of the proposed approach is presented in Fig.~\ref{fig:deep:dcf}. The figure illustrates long sequences (frames 0 to 200) and demonstrates the capability of individual tracking using spatial features extracted from the localisation model on both the UP-COUNT-TRACK and DroneCrowd datasets.

\subsubsection{Re-used deep visual features}\label{sec:deep_visual_features}

Originally, ECO combines features from the first (Conv-1) and last (Conv-5) convolutional layers of the VGG architecture~(\cite{chatfield2014return}) to generate feature representations for correlation filters, providing a broad range of information for general-purpose object tracking. However, these features are not well-suited for tracking tiny objects in drone-based crowd monitoring. First, the small size of objects, combined with the top-down drone perspective, necessitates both tailored feature adaptation and dense feature representation. Second, the high density of objects in the scene requires an efficient method for extracting deep features. The need of extracting features for each object separately makes the process of computing demanding.

\begin{figure}[h!]
\centering
\includegraphics[width=0.8\textwidth]{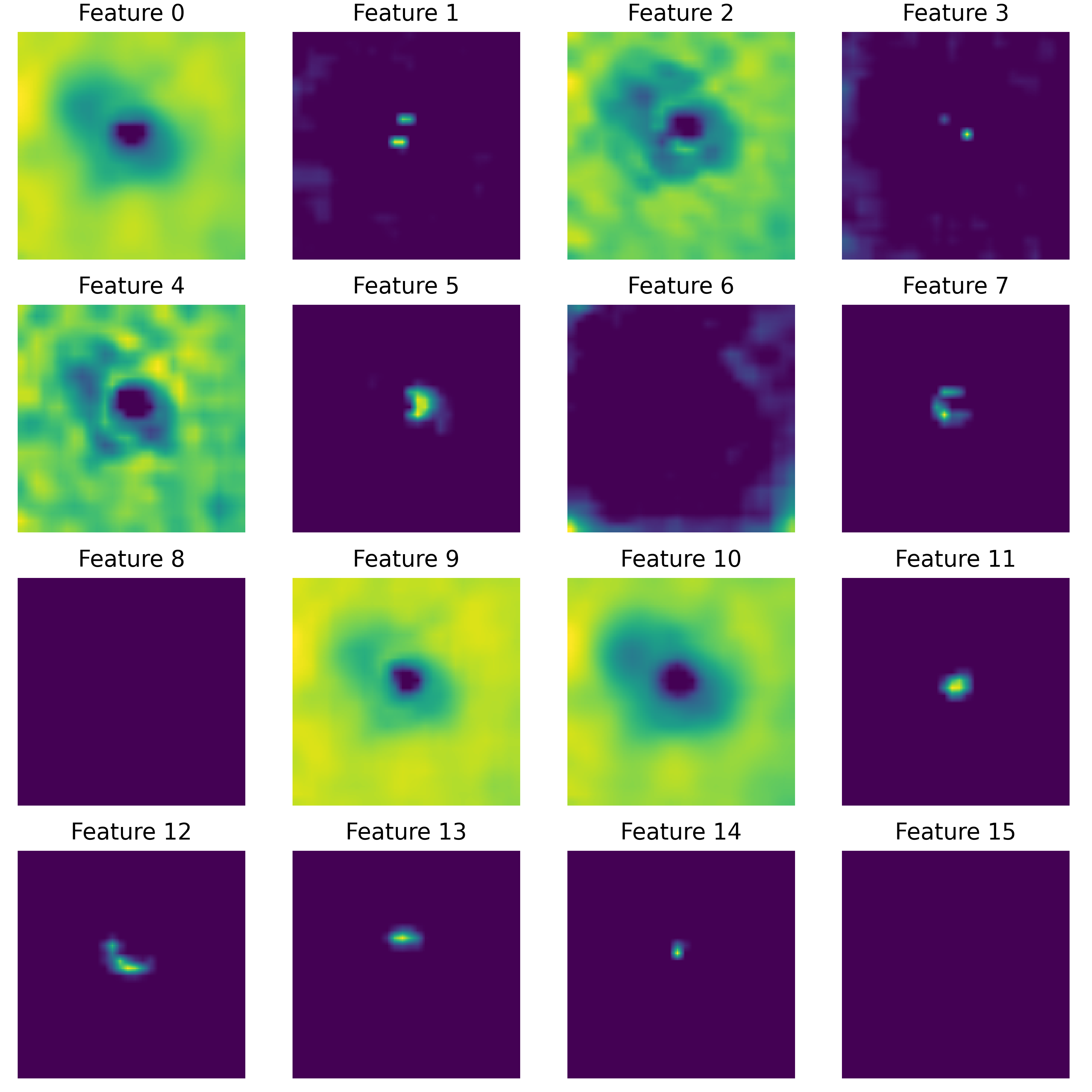}
\caption{Sixteen deep features extracted in the neighbourhood of a person. Values are normalised to a zero-one range for visualisation purposes.}
\label{fig:deep:features}
\end{figure} 

To address these challenges, we propose a solution aligned with the zero-waste machine learning paradigm~(\cite{trzcinski2024zero}). Instead of generating features through a dedicated neural network, we distill information directly from the people localisation method described in Section~\ref{sec:method:loc}. It employs a U-Net-like architecture consisting of an encoder for feature extraction, a decoder for spatial composition, and a head module that aggregates features and produces the final output. To maximise spatial resolution and feature richness, we extract features from the second-to-last layer of the model’s head, resulting in a spatial feature map of dimensions $544 \times 940 \times 16$. This approach generates features for the entire frame, which are then refined for individual objects by focusing on their surrounding regions. Additionally, these features extracted from the global map incorporate spatial and semantic context, providing more valuable information in densely populated scenes. An example visualisation of a feature set for a single object is illustrated in Fig.~\ref{fig:deep:features}. 

By repurposing features already computed for object localisation, the method aligns with the zero-waste machine learning paradigm by reducing computational overhead compared to employing a separate feature extraction network. This efficiency is needed for real-world drone-based applications, where computational resources are typically constrained.

\subsection{Tracking architecture integration}\label{sec:alltogether}

\begin{figure}[h!]
\centering
\includegraphics[width=\textwidth]{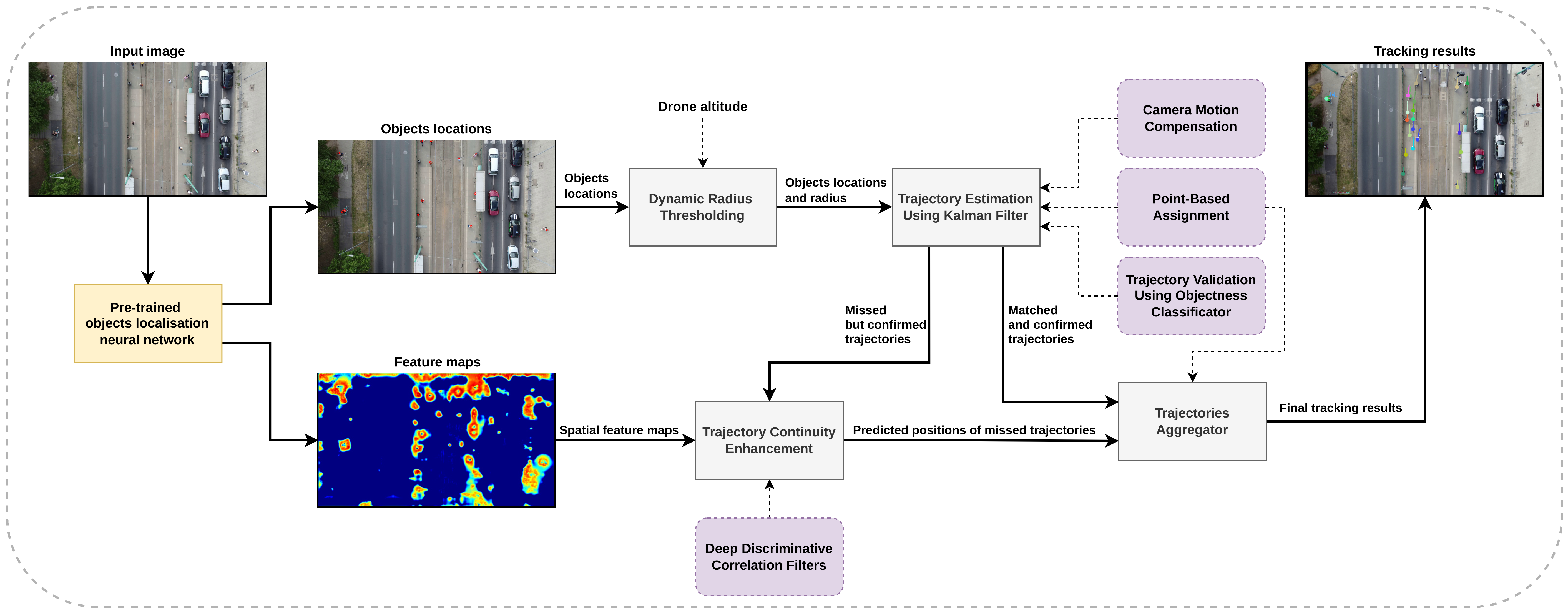}
\caption{Overview of the proposed tracking pipeline, integrating point-based object localisation, spatial feature maps and trajectory enhancement methods integration.}
\label{fig:tracking_pipeline}
\end{figure}

The previously discussed methodologies are integrated into a structured tracking system, forming a robust pipeline for drone-based crowd monitoring. The system operates sequentially, processing each video frame independently while maintaining temporal consistency across frames. Fig.~\ref{fig:tracking_pipeline} illustrates the overall architecture of the tracking algorithm. 

Initially, the localisation method detects individuals in each frame, providing coordinate-based outputs. Simultaneously, spatial feature maps are extracted from each frame and used as input for the trajectory enhancement module. Next, considering the drone's flight altitude, a dynamic threshold is calculated for each frame to align further processing with flight height. In the next step, a Kalman filter predicts the locations of known objects based on prior observations and corrects these predictions using a transformation matrix computed by the Camera Motion Compensation module. Using the estimated locations and calculated assignment threshold, newly detected individuals are matched to existing trajectories. This matching process employs the Hungarian algorithm and Euclidean distance metrics to associate coordinate sets. If a matched trajectory is not yet confirmed but has reached a sufficient age (lifespan), an additional classification step validates the trajectory to minimise missed detections. This process results in a set of matched and confirmed trajectories. When confirmed trajectories are not matched in a frame due to missed detections, their positions are estimated using Deep Discriminative Correlation Filters. By leveraging features extracted from the localisation network, this enhancement updates trajectories and assign them with Kalman filter estimations, improving continuity and smoothness. Finally, all trajectories are aggregated and returned using this method.

The integration of these components addresses key challenges in drone-based crowd monitoring:
\begin{itemize}
    \item Motion compensation corrects trajectory estimates by aligning object positions and mitigating the impact of camera movement noise.
    \item Dynamic assignment thresholds adapt to varying scene scales, enhancing tracking accuracy across different altitudes.
    \item Trajectory classification prior to confirmation addresses localisation algorithm imperfections by reducing false positives.
    \item The trajectory enhancement module maintains trajectory continuity despite temporary object occlusions or missed detections.
\end{itemize}

In the following section, we evaluate how these individual improvements contribute to overall performance gains, using two benchmark datasets.

\section{Evaluation}

\subsection{Evaluation datasets}

In our experiments, we utilise the newly proposed UP-COUNT-TRACK dataset (Section~\ref{sec:dataset}) and the publicly available DroneCrowd dataset~(\cite{wen2021detection}). DroneCrowd is the first large-scale dataset specifically designed for localising point-oriented individuals in UAV-recorded videos. It includes 112 video sequences captured across various environments, with 30 sequences dedicated to testing. The test subset contains 9,000 frames, evenly distributed with 300 frames per sequence. Although the number of trajectories per sequence ranges from 44 to 296, the dataset has a higher average of 169.7, resulting in a total of 5,092 unique trajectories competing with UP-COUNT-TRACK. Despite the stationary nature of the drone recordings and short video durations, the dataset poses challenges due to the tiny size of the objects. Detailed dataset comparison is included in Table~\ref{tab:dataset:comparsion2}.

\begin{table}[ht!]
\centering
\caption{Statistical comparison of the DroneCrowd and UP-COUNT-TRACK datasets for the people trajectory counting task (considering only test subsets).}
\label{tab:dataset:comparsion2}
\resizebox{0.8\textwidth}{!}{%
\begin{tabular}{|c|cc|}
\hline
\multirow{2}{*}{\textbf{Trajectory counting statistics}} & \multicolumn{2}{c|}{\textbf{\begin{tabular}[c]{@{}c@{}}Dataset name\\ (only test subset)\end{tabular}}} \\ \cline{2-3} 
                                                         & \multicolumn{1}{c|}{\textbf{DroneCrowd}}                    & \textbf{UP-COUNT-TRACK}                   \\ \hline
Number of sequences                                      & \multicolumn{1}{c|}{30}                                     & 31                                        \\ \hline
Min. frames number                                       & \multicolumn{1}{c|}{300}                                    & 361                                       \\ \hline
Mean frames number                                       & \multicolumn{1}{c|}{300}                                    & 1088.7                                    \\ \hline
Max. frames number                                       & \multicolumn{1}{c|}{300}                                    & 2,541                                     \\ \hline
Total frames                                             & \multicolumn{1}{c|}{9,000}                                  & 33,751                                    \\ \hline
Min. trajectories                                        & \multicolumn{1}{c|}{44}                                     & 13                                        \\ \hline
Mean trajectories                                        & \multicolumn{1}{c|}{169.7}                                  & 122.8                                     \\ \hline
Max. trajectories                                        & \multicolumn{1}{c|}{296}                                    & 1182                                      \\ \hline
Total trajectories                                       & \multicolumn{1}{c|}{5,092}                                  & 3,807                                     \\ \hline
\end{tabular}%
}
\end{table}

\subsection{Evaluation metrics}

\textbf{Higher Order Tracking Accuracy (HOTA)}~(\cite{luiten2021hota}) metric assesses multi-object tracking performance by simultaneously evaluating detection and association accuracy, offering a more nuanced assessment of tracking algorithms. HOTA addresses the limitations of traditional metrics like MOTA (Multiple Object Tracking Accuracy)~(\cite{bernardin2008evaluating}) and ID-F1 (metric of the consistency of identity tracking across a sequence) by integrating them into a unified framework that considers detection, association, and localisation accuracy collectively.

\textbf{Tracking mean Average Precision (T-mAP)} is applied to evaluate the tracking precision of point-oriented methods. It is determined by the distance threshold calculated using the greedy method, according to the procedure introduced in~(\cite{wen2021detection}). Tracklets, grouped by identity and ranked by average detection confidence, are considered correct if they match ground-truth tracklets above a specified threshold. The results are also reported for the selected distance thresholds of 10 pixels (T-AP@10), along with T-mAP, which provides cumulative results in thresholds ranging from 1 to 25 pixels.

\textbf{Identification Switches (ID-SW)}~(\cite{5206735}) metric measures how often a tracker incorrectly reassigns an object's ID, resulting in a break in the object's trajectory. This typically occurs due to misassignment or a loss of detection continuity. The ID-SW metric is essential for assessing the consistency of a tracking algorithm in maintaining correct object identities over time. The score is calculated by counting the total number of identity switches across all sequences and then averaging them.

\textbf{Tracking Mean Absolute Error (Tr-MAE)} is a metric used to evaluate counting errors for object trajectories within each sequence in a dataset. It is specifically designed to validate tracking methods by assessing an algorithm's ability to count objects effectively while considering object identifications across frames. The metric is defined as:
\begin{equation}
    Tr \text{-} MAE = \frac{1}{n} \sum_{i=1}^{n} |y_i - \hat{y}_i|
\end{equation}
where $y$ represents the ground truth number of trajectories, and $\hat{y}$ denotes the estimated unique object's count.

\textbf{Tracking Normalised Mean Absolute Error (Tr-nMAE)} calculates the relative counting error, focusing on unique trajectories within a sequence. It can be interpreted as the percentage counting error for the entire sequence (video), providing a measure of an algorithm's performance in tracking and counting objects. It is defined as:
\begin{equation}
    Tr \text{-} nMAE = \frac{1}{n} \sum_{i=1}^{n} \frac{|y_i - \hat{y}_i|}{y_i}
\end{equation}
where $y$ represents the ground truth trajectory count, and $\hat{y}$ denotes the estimated trajectory count for a sequence.

\subsection{Tracking results}

Tables \ref{tab:my-table} and \ref{tab:results:dronecrowd} present the tracking performance on the UP-COUNT-TRACK and DroneCrowd datasets, respectively, evaluating the impact of proposed improvements. We compare a baseline tracking method with incremental additions of Camera Motion Compensation (CMC), dynamic thresholding considering flight altitude (ALT), additional classification in the trajectory confirmation step (CLS), and the enhanced Deep Discriminative Correlation Filters (DDCF). Additionally, we evaluate our method against the state-of-the-art, globally optimal greedy (GOG) offline approach, demonstrating the robustness of our online tracking method across different datasets and challenging conditions, even when compared to an offline solution.

\begin{table}[ht!]
\centering
\caption{Tracking results for the UP-COUNT-TRACK dataset, considering proposed improvements. *Globally-optimal greedy (GOG) algorithm is an offline method.}
\label{tab:my-table}
\resizebox{\textwidth}{!}{%
\begin{tabular}{|l|c|c|c|c|c|c|}
\hline
\textbf{Method}                                                                  & \textbf{HOTA ($\uparrow$)} & \textbf{T-mAP ($\uparrow$)} & \textbf{T-AP@10 ($\uparrow$)} & \textbf{ID-SW ($\downarrow$)} & \textbf{Tr-MAE ($\downarrow$)} & \textbf{Tr-nMAE ($\downarrow$)} \\ \hline
Baseline                                                                         & 0.63                      & 37.04                      & 39.34                        & 3305                         & $49.13 \pm 117.22$            & $0.37 \pm 0.34$                \\ \hline
\begin{tabular}[c]{@{}l@{}}Baseline\\ + CMC\end{tabular}                         & 0.63                      & 37.36                      & 39.59                        & 3180                         & $48.84 \pm 114.43$            & $0.38 \pm 0.32$                \\ \hline
\begin{tabular}[c]{@{}l@{}}Baseline\\ + CMC\\ + ALT\end{tabular}                 & 0.64                      & 38.68                      & 40.90                        & 3126                         & $44.97 \pm 103.25$            & $0.33 \pm 0.29$                \\ \hline
\begin{tabular}[c]{@{}l@{}}Baseline\\ + CMC\\ + ALT\\ + CLS\end{tabular}         & 0.63                      & 38.72                      & 40.95                        & 2943                         & $41.81 \pm 96.76$             & $0.30 \pm 0.26$                \\ \hline
\begin{tabular}[c]{@{}l@{}}Baseline\\ + CMC\\ + ALT\\ + CLS\\ + DDCF\end{tabular} & \textbf{0.63}                      & \textbf{44.35}                      & \textbf{45.88}                        & \textbf{287}                          & \textbf{20.45} $\pm$ \textbf{44.81}             & \textbf{0.15} $\pm$ \textbf{0.11}                \\ \hline \hline
GOG*                                                                             & 0.42                      & 36.21                      & 37.63                        & 1868                         & $64.19 \pm 110.24$            & $0.57 \pm 0.36$                \\ \hline
\end{tabular}%
}
\end{table}

\begin{table}[ht!]
\centering
\caption{Tracking results for the DroneCrowd dataset, considering proposed improvements. *Globally-optimal greedy (GOG) algorithm is an offline method.}
\label{tab:results:dronecrowd}
\resizebox{\textwidth}{!}{%
\begin{tabular}{|l|c|c|c|c|c|c|}
\hline
\textbf{Method}                                                                  & \textbf{HOTA ($\uparrow$)} & \textbf{T-mAP ($\uparrow$)} & \textbf{T-AP@10 ($\uparrow$)} & \textbf{ID-SW ($\downarrow$)} & \textbf{Tr-MAE ($\downarrow$)} & \textbf{Tr-nMAE ($\downarrow$)} \\ \hline
Baseline                                                                         & 0.53                      & 46.27                      & 49.99                        & 6290                         & $57.47 \pm 37.13$             & $0.32 \pm 0.15$                \\ \hline
\begin{tabular}[c]{@{}l@{}}Baseline\\ + CMC\end{tabular}                         & 0.53                      & 46.87                      & 50.37                        & 6231                         & $55.83 \pm 35.53$             & $0.31 \pm 0.14$                \\ \hline
\begin{tabular}[c]{@{}l@{}}Baseline\\ + CMC\\ + ALT\end{tabular}                 & 0.52                      & 47.44                      & 51.13                        & 6771                         & $52.90 \pm 31.37$             & $0.30 \pm 0.13$                \\ \hline
\begin{tabular}[c]{@{}l@{}}Baseline\\ + CMC\\ + ALT\\ + CLS\end{tabular}         & 0.52                      & 47.32                      & 50.88                        & 6645                         & $52.10 \pm 30.36$             & $0.30 \pm 0.14$                \\ \hline
\begin{tabular}[c]{@{}l@{}}Baseline\\ + CMC\\ + ALT\\ + CLS\\ + DDCF\end{tabular} & \textbf{0.54}                      & \textbf{54.59}                      & \textbf{57.04}                        & \textbf{388}                          & \textbf{37.60} $\pm$ \textbf{25.78}             & \textbf{0.23} $\pm$ \textbf{0.16}                \\ \hline \hline
GOG*                                                                             & 0.51                      & 50.47                      & 53.21                        & 1326                         & 38.10 $\pm$ 31.14             & 0.24 $\pm$ 0.17                \\ \hline
\end{tabular}%
}
\end{table}

Overall, incrementally added improvements enhance tracking and counting metrics for both datasets. Notably, trajectory continuity enhancement using DDCF has a significant impact, particularly on reducing ID switches and improving the trajectory counting metric. On UP-COUNT-TRACK (Table \ref{tab:my-table}), the fully enhanced method achieves the best performance across all metrics, reaching a HOTA score of 0.63, T-mAP of 44.35, while drastically reducing ID switches to 287 and trajectory counting error to 15\%. Similarly, on DroneCrowd (Table \ref{tab:results:dronecrowd}), the fully improved method achieves the highest HOTA (0.54) and T-mAP (54.59), with ID switches reduced to 388 and trajectory counting error to 23\%. 

While our method achieves tracking metrics comparable to GOG on DroneCrowd, it still demonstrates an advantage in all other metrics, particularly in reducing ID switches. In contrast, on UP-COUNT-TRACK, our enhanced online method significantly outperforms the offline GOG algorithm. This difference may stem from more significant visual changes between frames and longer sequences, which pose tracking challenges without trajectory continuity improvements. These results consistently validate the effectiveness of our proposed online tracking approach across different datasets and challenging scenarios, including drone-view crowd tracking.

\begin{figure}[h!]
\centering
\includegraphics[width=0.95\textwidth]{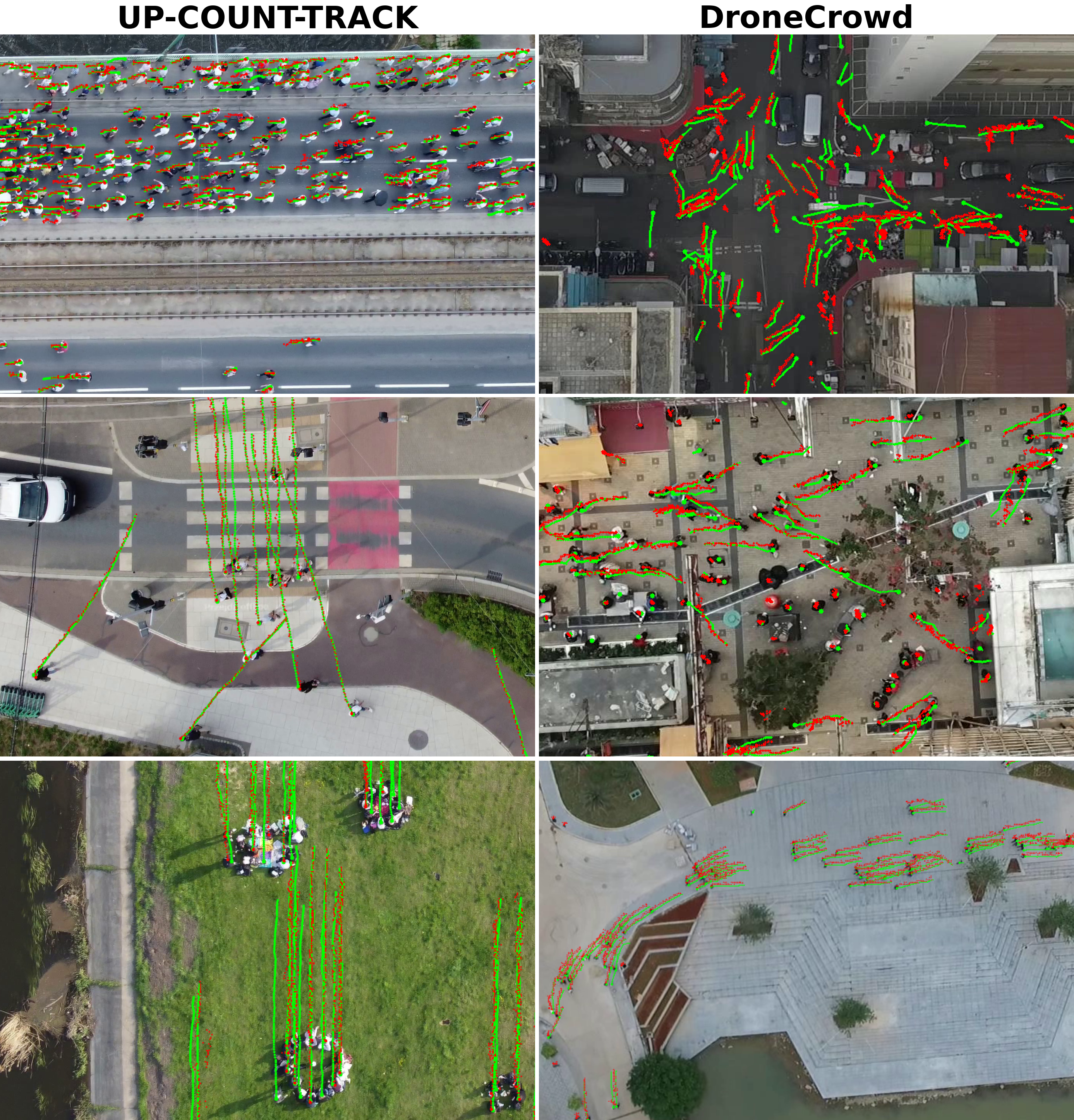}
\caption{The visual comparison of ground-truth trajectories (green) and estimated results (red) for both datasets.}
\label{fig:results:examples}
\end{figure}

To visualise tracking results on both datasets, Fig.~\ref{fig:results:examples} presents examples from various environments and flight altitudes. Ground-truth trajectories are marked with green dots, while the model’s predicted trajectories are shown in red. This visualisation illustrates the accuracy and continuity of the estimations.

\subsection{Statistical analysis}

\begin{figure}[h!]
\centering
\includegraphics[width=\textwidth]{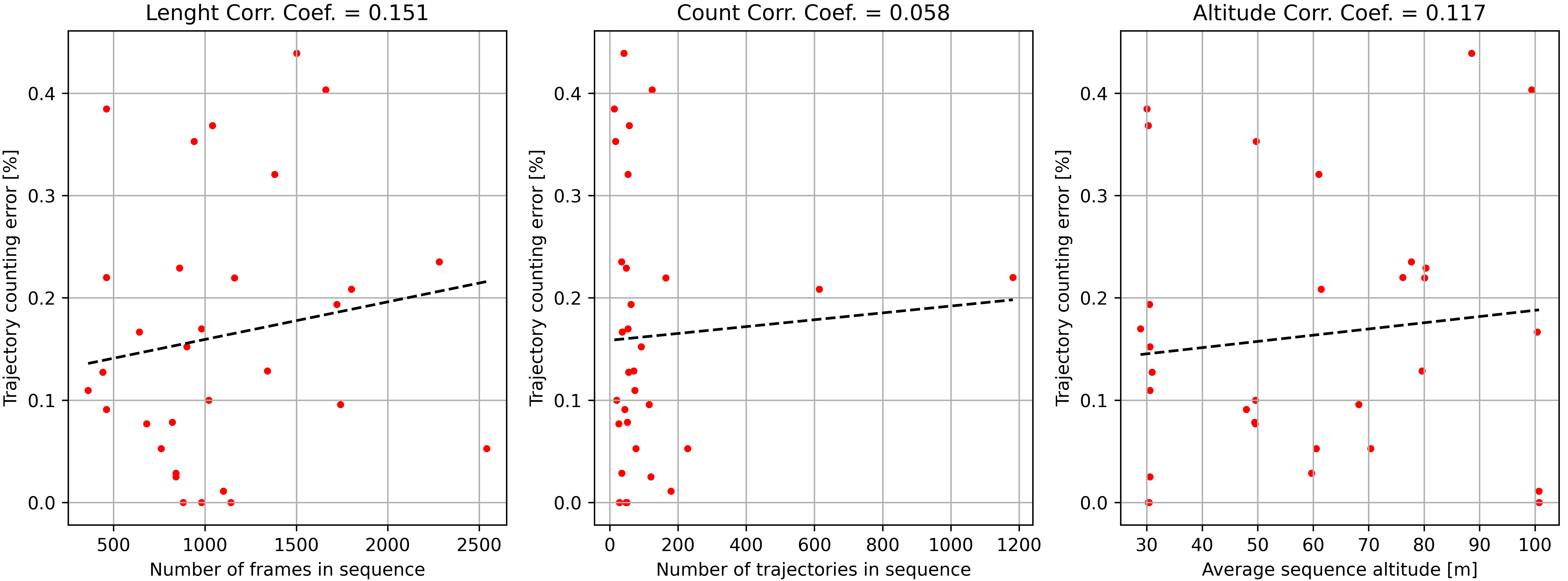}
\caption{Statistical analysis of counting trajectory error and three sequence characteristics: sequence length, number of unique trajectories in a sequence, and average flight altitude during recording.}
\label{fig:ablation:stats}
\end{figure}

To evaluate the algorithm's performance under varying conditions, a statistical analysis was conducted using the UP-COUNT-TRACK dataset (Fig.~\ref{fig:ablation:stats}). The study examined the Pearson correlation between trajectory counting errors (Tr-nMAE) and three sequence characteristics: sequence length, the number of unique trajectories, and average flight altitude. The calculated correlation coefficients for these characteristics are 0.151, 0.058, and 0.117, respectively, indicating negligible correlation. These results suggest that the algorithm's performance is largely unaffected by variations in tracking duration, objects' density, or changes in objects' sizes.

\subsection{Classification model results}\label{sec:cls:results}

\begin{figure}[h!]
\centering
\includegraphics[width=0.9\textwidth]{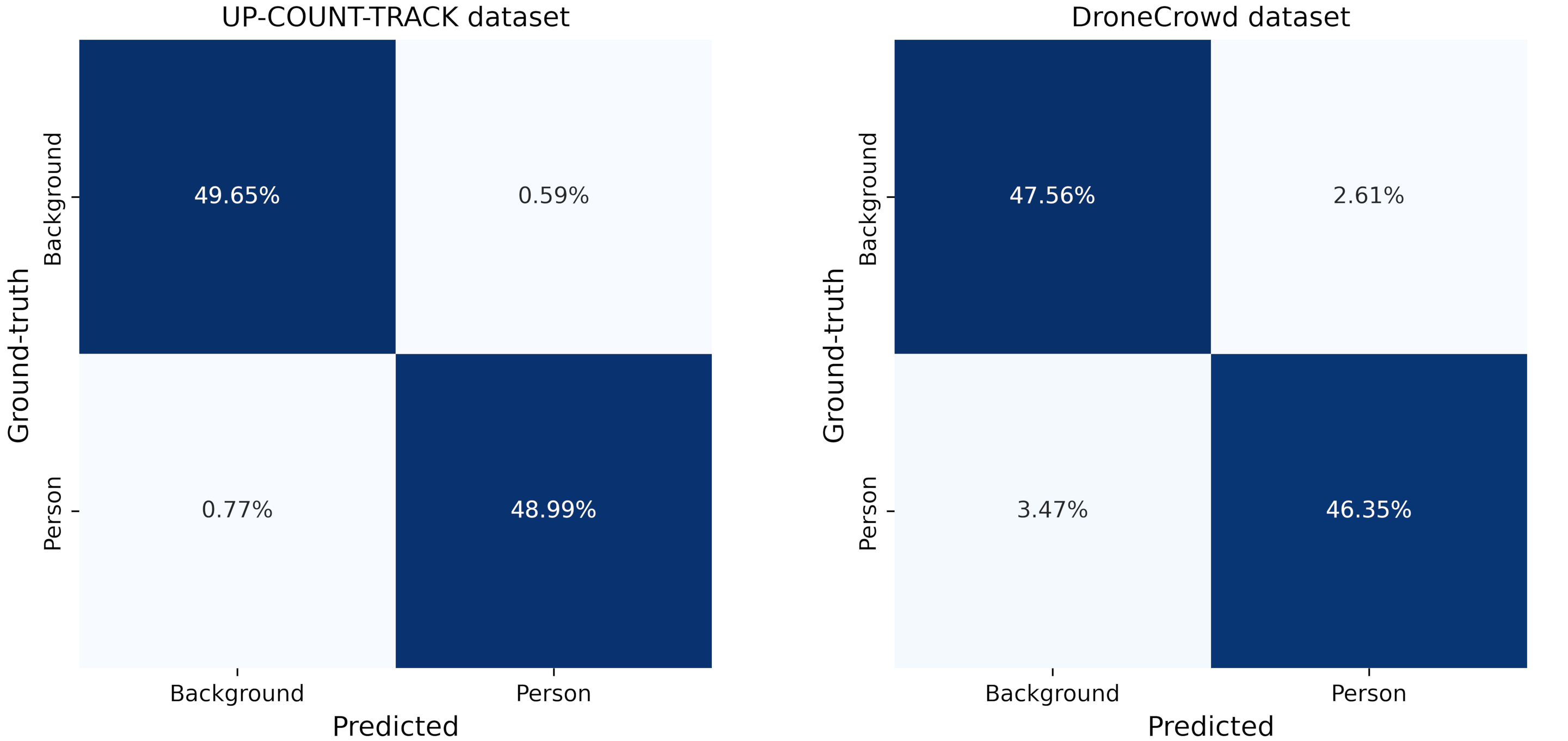}
\caption{Confusion matrices of the classification algorithm for the UP-COUNT-TRACK and DroneCrowd datasets.}
\label{fig:cls:confusion_matrix}
\end{figure}

Due to the balanced data, the accuracy metric can be used to evaluate the classification algorithm. For the UP-COUNT-TRACK dataset, the accuracy for the background and person classes is 0.988 and 0.985, respectively, resulting in an overall accuracy of 0.987. For the DroneCrowd dataset, the corresponding accuracies are 0.948 and 0.930, with a total accuracy of 0.939 for both classes. Fig.~\ref{fig:cls:confusion_matrix} presents the confusion matrices for both datasets, illustrating the algorithm's effectiveness in distinguishing between background and person classes.

\section{Potential applications}

\begin{figure}[ht!]
\centering

     \begin{subfigure}[b]{0.95\textwidth}
         \centering
         \includegraphics[width=\textwidth]{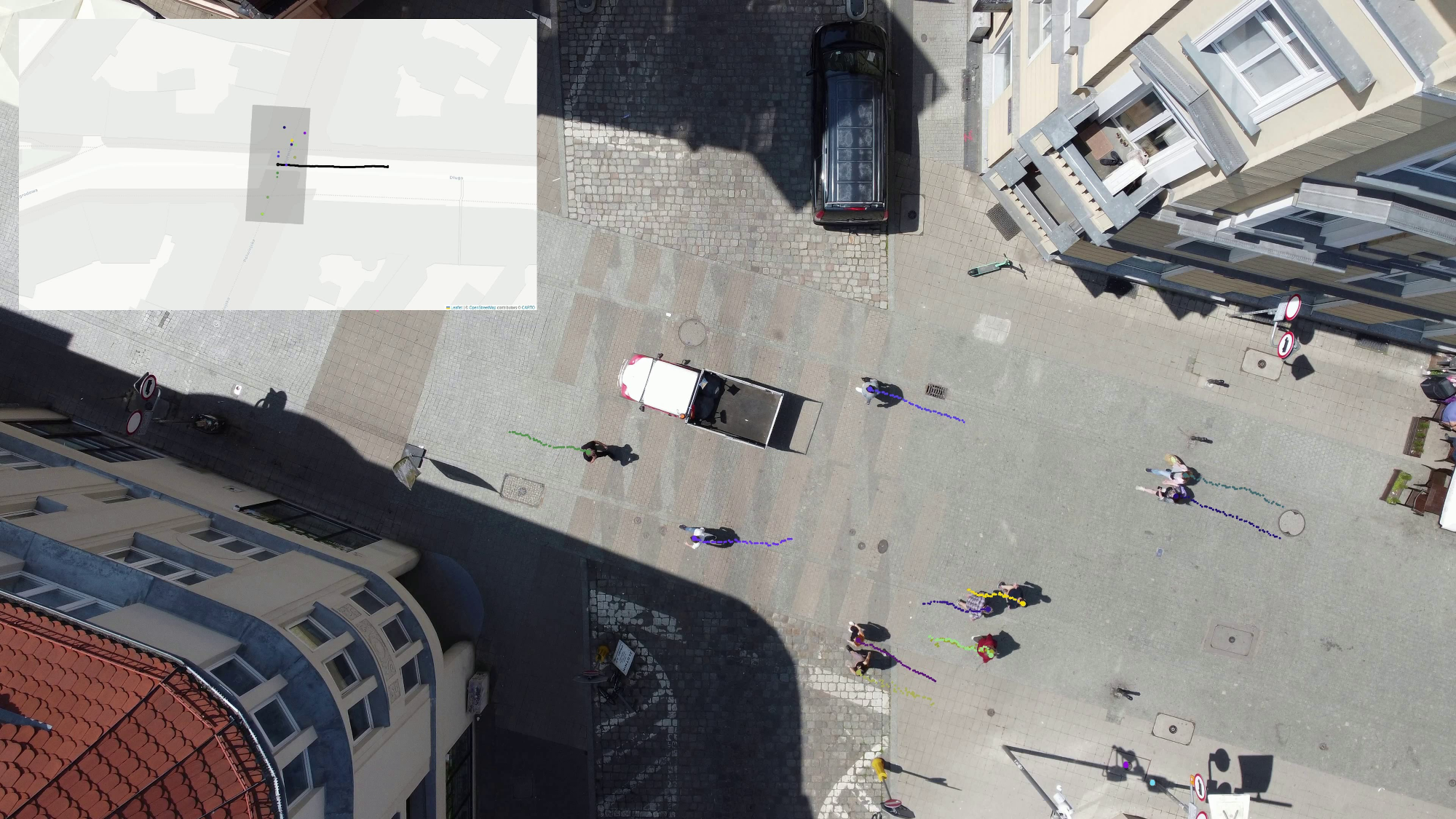}
         \caption{Example frame from video recorded at low flight altitude. }
     \end{subfigure}
     
     \begin{subfigure}[b]{0.95\textwidth}
         \centering
         \includegraphics[width=\textwidth]{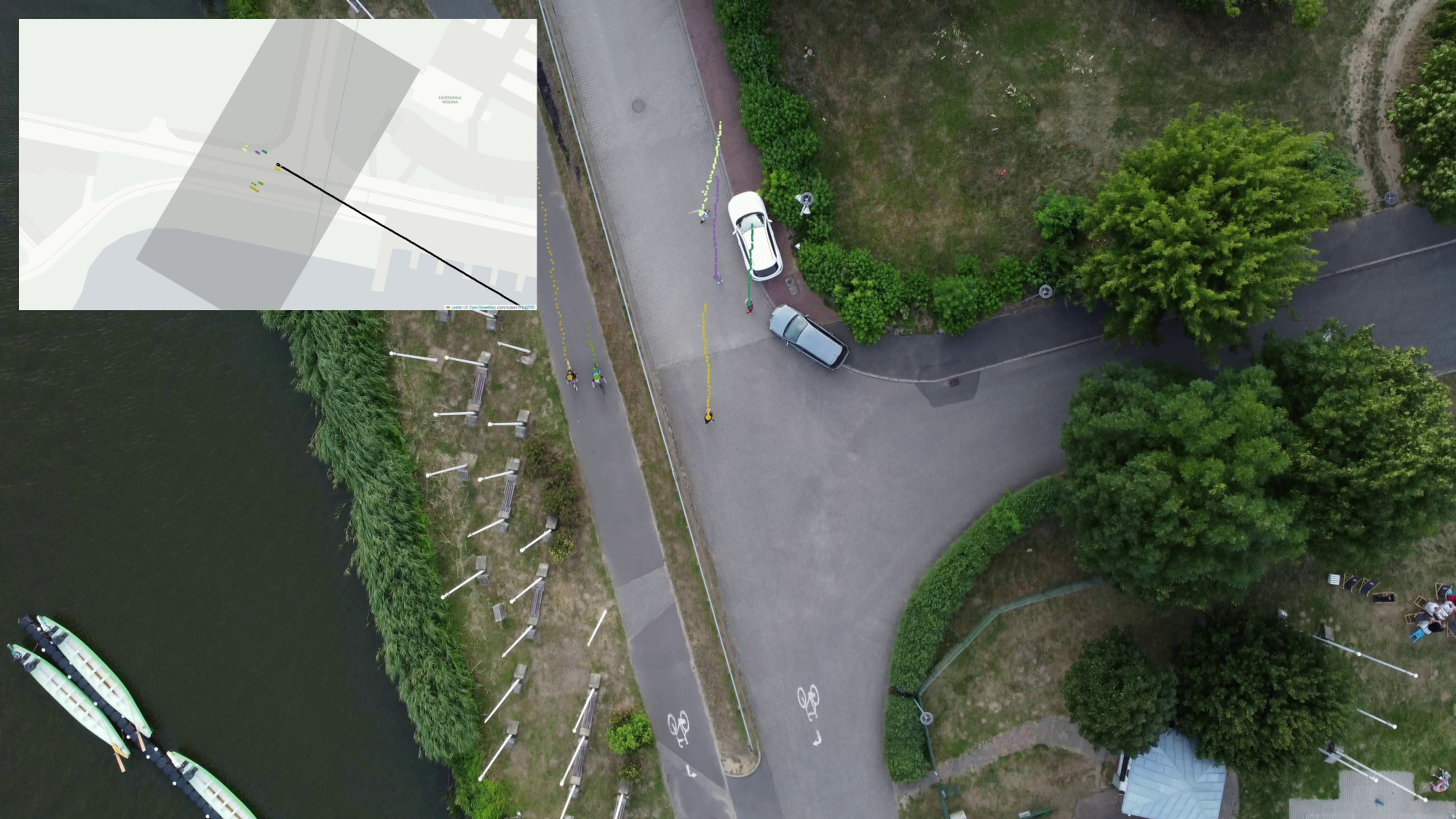}
         \caption{Example frame from video recorded at medium flight altitude.}
     \end{subfigure}
     
\caption{Example screenshots from a visualisation application that combines people's trajectories with drone sensors to estimate their coordinates, generating a map (left top). In-frame tracking information is merged with the position and the interior drone's state to provide accurate object mapping. The gray area on the map is the field of view of the drone's camera, while the black line is the drone's position history.}
\label{fig:flowanalysis}
\end{figure}

The proposed algorithm enhances tracking of individuals in drone-recorded videos, enabling a detailed analysis of movement patterns. This technology extends beyond crowd size estimation and identification in high-density areas to include monitoring crowd dynamics, capturing movement trajectories, and analysing behavioural patterns. Such insights support data-driven urban planning, optimising public spaces to improve pedestrian flow, reduce congestion, and identify areas requiring infrastructure enhancements. This is achieved by integrating the algorithm’s results with drone sensor data, including global coordinates, north heading, and altitude above ground level, ensuring accurate mapping locations of individuals onto a global reference frame. Based on these principles, a proof-of-concept application has been developed. Example screenshots from the application are presented in Fig.~\ref{fig:flowanalysis}, showing side-by-side comparisons of the processed video and the corresponding map. These visuals highlight detected objects and their movement histories, demonstrating the system’s ability to track individuals over time. The proposed system is not limited to people's localisation and tracking. With the addition of an appropriately prepared dataset, it can be easily adapted to operate with any object type.

\section{Conclusions}


This paper presents a deep-learning-based system that integrates modified Deep Discriminative Correlation Filters with cost-effective trajectory refinement techniques to ensure robust individual tracking in drone-based video feeds. The system dynamically adapts to environmental conditions, improving tracking reliability. Its effectiveness was evaluated on the DroneCrowd dataset and the newly introduced UP-COUNT-TRACK dataset, achieving counting errors of 23\% and 15\%, respectively. Furthermore, the publicly available UP-COUNT-TRACK dataset serves as a valuable resource for benchmarking and advancing crowd analysis algorithms, enhancing the potential of the proposed system for accurate and efficient crowd monitoring in complex real-world scenarios.

The proposed method extends beyond conventional surveillance and security applications. Monitoring and analysing crowd dynamics in urban environments is essential for public safety, emergency response, and infrastructure planning. Drone-based crowd monitoring provides advantages over static surveillance systems by enabling the observation of large-scale gatherings, traffic flows, and pedestrian movement patterns. In emergency response scenarios, real-time pedestrian tracking can support evacuation planning, crowd dispersion analysis, and rapid threat detection. Additionally, urban planners can utilise trajectory analytics to optimise public transportation networks, pedestrian pathways, and smart city infrastructure. The integration of deep learning-based tracking in drone platforms enhances intelligent monitoring systems, contributing to automated, adaptive, and real-time decision-making in various domains.

\subsection{Limitations and future directions}

Although the method employs an online tracking approach, it remains computationally demanding, limiting its feasibility for onboard drone deployment and real-time processing. Addressing these challenges requires further research and optimisation, including the development of more efficient algorithms, hardware-accelerated implementations, and model simplification techniques to enhance performance while maintaining tracking accuracy.

\section*{Ethics statement}
The UP-COUNT-TRACK dataset maintains data privacy. The aerial perspective captures individuals at a resolution where identifiable features are minimised, reducing the risk of personal identification and mitigating privacy concerns. The dataset does not include biometric data, facial features, or personally identifiable information, ensuring compliance with ethical standards for data collection and usage.

While this research focuses on civilian applications such as crowd monitoring for public safety and crowd management, it is essential to acknowledge that the method could potentially be adapted for applications in surveillance, target identification and tracking, highlighting the importance of responsible development and usage.

\section*{Data availability statement}

The data supporting the findings of this research are publicly available at \url{https://doi.org/10.5281/zenodo.13829572}.

\bibliography{references}  


\end{document}